%% file: 0_eccv2022submission.tex
\documentclass[runningheads]{llncs}
\usepackage{graphicx}

\usepackage{tikz}
\usepackage{comment}
\usepackage{amsmath,amssymb} 
\usepackage{color}
\usepackage{authblk}
\usepackage{hyperref}
\usepackage{url}
\usepackage{graphicx}
\graphicspath{{./figures/}}
\usepackage{tabularx}
\usepackage{booktabs} 
\usepackage{caption}
\usepackage{subcaption}
\usepackage{multirow}
\usepackage[normalem]{ulem}
\usepackage{soul}
\usepackage{algorithm}
\usepackage{array}
\usepackage[noend]{algpseudocode}
\usepackage{pifont}
\usepackage[accsupp]{axessibility}  

\begin{document}
\pagestyle{headings}
\mainmatter
\def\ECCVSubNumber{7693}  

\title{Federated Self-supervised Learning\\ for Video Understanding} 

%

\author{Yasar Abbas Ur Rehman\inst{1,}\thanks{Equal contribution, authors ordered alphabetically.} \and
Yan Gao\inst{2,\star} \and
Jiajun Shen\inst{1} \and Pedro Porto Buarque de Gusmão\inst{2} \and Nicholas Lane\inst{2,3}}

\authorrunning{Y. Rehman and Y. Gao et al.}
%
\institute{TCL AI Lab, Hong Kong,  \and
University of Cambridge, United Kingdom, 
\and Samsung AI Cambridge, United Kingdom}


\maketitle

\begin{abstract}
The ubiquity of camera-enabled mobile devices has lead to large amounts of unlabelled video data being produced at the edge. Although various self-supervised learning (SSL) methods have been proposed to harvest their latent spatio-temporal representations for task-specific training, practical challenges including privacy concerns and communication costs prevent SSL from being deployed at large scales. To mitigate these issues, we propose the use of Federated Learning (FL) to the task of video SSL. In this work, we evaluate the performance of current state-of-the-art (SOTA) video-SSL techniques and identify their shortcomings when integrated into the large-scale FL setting simulated with kinetics-400 dataset. We follow by proposing a novel federated SSL framework for video, dubbed FedVSSL, that integrates different aggregation strategies and partial weight updating.   
Extensive experiments demonstrate the effectiveness and significance of FedVSSL as it outperforms the centralized SOTA for the downstream retrieval task  by $6.66\%$ on UCF-101 and $5.13\%$ on HMDB-51.

\end{abstract}

\input{1_introduction}

\input{2_related_work}
\input{3_method}

\input{4_evaluation}

\input{5_conclusions}

\clearpage
%
%
\bibliographystyle{splncs04}
\bibliography{egbib}
\end{document}

%% file: 1_introduction.tex
\section{Introduction}

A plethora of video content, often unlabelled and of private nature, is generated everyday from  cameras in cellphones, tablets, and other mobile devices \cite{goyal2022vision,kay2017kinetics,park2020sinet,aytar2016soundnet,vondrick2016anticipating}.
Being able to repurpose this data to solve various tasks in computer vision  has been of great interest to researchers since the last decade \cite{chen2020simple,he2020momentum,kolesnikov2019revisiting,doersch2015unsupervised,piergiovanni2020evolving,wang2015unsupervised,wang2020self,goyal2022vision}. Self-Supervised Learning (SSL) allows us to harvest these data contents by learning \textit{intermediate} visual representations from unlabelled data, which can then be used as a starting point to solve specific downstream tasks (e.g., human action recognition \cite{soomro2012ucf101}, temporal action detection \cite{zhao2017temporal}). In practice, however, deploying SSL in its na\"ive form would require massive amounts of data to be sent to a centralized server for processing, posing significant concerns around privacy, \cite{jain2021biometrics,hu2022dynamic}, communication, and storage costs, ultimately limiting the technology to small datasets.

A natural way to mitigate those issues is to combine SSL with the novel decentralized machine learning technique known as Federated Learning (FL)  \cite{mcmahan2017communication}. In FL, distributed population of edge devices collaboratively train a shared model while keeping their personal data private. Essentially, FL dilutes the burden of training across devices and avoids privacy and storage issues by not collecting users' data samples.  The potential integration of video-SSL and FL into one coherent system offers lots of benefits in addition to data privacy. It enables large-scale decentralized feature learning from real-world data without requiring any costly and laborious data annotations. This can practically improve the performance and enables a vast majority of vision models for video applications \cite{goyal2022vision}, which have been under the shadow otherwise. Surprisingly, bearing such potential benefits there is no prior work that has so far studied video-SSL in FL.

In this paper, we conduct the first comprehensive study on video-SSL training in \textit{cross-device} FL environment \cite{kairouz2019advances}. Our key findings from this study shows that: (1) The vanilla FL pretraining of video-SSL approaches are surprisingly not affected by the distribution of the data either being IID or non-IID. (2) Video-SSL with FL performs significantly better than the corresponding centralized SSL pretraining on video retrieval tasks and comparatively worse when the model is fined-tuned for action recognition. (3) We also show that the video SSL methods in FL settings are computationally efficient, regularized, and resilient to small-scale perturbation, compared to their centralized counterparts.

Based on the findings of our study of video-SSL with FL, we propose a novel federated learning framework, \textit{\textbf{FedVSSL}},  designed specifically for video SSL. This framework allows to transceive only the backbone parameters of the video-SSL model during each communication round in FL. It then leverages a novel aggregation strategy, inspired by stochastic weighted averaging (SWA) \cite{izmailov2018averaging}, to aggregate and update the weights of the clients (performing video-SSL) at the server. FedVSSL obtains the state-of-the-art (SOTA) performance in video clip retrieval and competitive performance on action recognition against FedAvg and centralized video-SSL. The main contributions of this work are as follows:

\begin{enumerate}
    \item We  conduct the first systematic study of training video-SSL methods in FL \textit{cross-device} settings with a large number of distributed clients. This establishes a baseline for na\"ively implementing various video-SSL techniques using FL; shedding light on the basic problems of integrating video-SSL with FL into one coherent system.
    \item Based on the above, we propose a general FL framework, FedVSSL, based on SWA \cite{izmailov2018averaging} for pretraining video-SSL methods in FL. Our method obtains SOTA performance in video clip retrieval and competitive performance on action recognition against FedAvg and centralized video-SSL.
    \item We release our code and models on GitHub \footnote{\url{https://github.com/yasar-rehman/FEDVSSL}} to allow for reproducibility and stimulate further research in the field. 
\end{enumerate}

%% file: 2_related_work.tex
\section{Background and Related Work}
\subsection{Video Self-supervised Representation Learning}
Video-SSL approaches often rely on solving a pretext task~\cite{feichtenhofer2021large} in an unsupervised fashion to learn representations that can be reused in solving other downstream tasks. The pretext tasks in video-SSL are either based on contrastive methods, non-contrastive methods, or a combination of both. Contrastive methods exploit the similarity between the two augmented views of the input samples using contrastive learning to learn the spatio-temporal representation from the unlabeled data \cite{han2020self,han2019video,li2020learning,feichtenhofer2021large,romijnders2021representation}. On the other hand, non-contrastive approaches utilize specialized pretext tasks to generate pseudo signals to learn the spatio-temporal representation in a supervised manner, usually requiring just a single input \cite{jing2018self,misra2016shuffle,xu2019self,jenni2020video,benaim2020speednet,yao2020video,wang2020self,wang2021unsupervised,lee2017unsupervised}. 

Regardless of the specific SSL method being deployed, a recent study has shown that SSL models become more robust on a wide range of vision-based tasks when pretrained on real-world and uncurated data \cite{goyal2022vision}. Bearing such tremendous potential, video-SSL models can extend the horizons of the many vision-based applications. However, the utility of these video-SSL models is significantly limited by the scale of the datasets available on the training server due to issues such as data privacy, communication cost and large data storage requirements.
In this paper, we extend the utility of video-SSL methods beyond the centralized servers. This would allow harvesting of information from an unprecedented amount of user data, offering new opportunities to advance the quality and robustness of video-SSL models.

\subsection{Federated Visual Representation Learning}
Federated Learning (FL) \cite{mcmahan2017communication} has received a lot of attention in recent years.  In this new paradigm, the server now needs to \emph{aggregate} incoming weights from clients to progressively produce better networks. In its original form \cite{mcmahan2017communication}, aggregation was performed using FedAvg, an \emph{aggregation strategy} that generates a model via a weighted sum of clients' parameters. However, this simple aggregation method can perform particularly poorly in realistic scenarios where clients have very different data distributions, i.\ e.\ the available datasets are essentially not Independent and Identically Distributed (IID).

In the past few years, a number of aggregation strategies have been proposed to improve upon the original FedAvg. Authors in \cite{gao2021end} introduce local-model training loss as a weighting coefficient for aggregation. Adaptive federated optimization approaches, proposed in \cite{reddi2020adaptive}, incorporate knowledge of past iterations by applying a separate gradient-based optimization on the server-side. Specific to SSL, authors in \cite{zhang2020federated} suggest using contrastive-learning on the server-side based on the logits sent by the clients, thus imposing some privacy issues.

Besides choosing a specific aggregation method when training using SSL in FL, we must also decide which parts of the networks need to be aggregated as some of the weights are associated with latent representations (backbone) while others are simply used for solving the pretext tasks (head). To this end, authors in \cite{zhuang2021collaborative} proposed FedU, that determines the update of the clients' model classification head based on the degree of backbone model divergence between the server and the clients. 

It is important to note that the above-mentioned works were applied to small-scale \textit{image-based} datasets \cite{krizhevsky2009learning,russakovsky2015imagenet} and that their proposed aggregation methods \emph{rely on the usual class-label based definition of non-IID}. However, SSL should not depend on class-based labels, which prompts the question of whether such a definition of non-IID datasets bears any impact in video-SSL in FL. To the best of our knowledge, ours is the first FL SSL training framework tailored to \emph{video} that is capable of producing SOTA models better than their centralized counterparts, regardless of the pretext class-label non-IID partitioning. We do this by correctly selecting only the backbone weights for aggregation.

%% file: 3_method.tex
\section{Methodology}
In this section, we provide the details of our systematic study on video-SSL using Federated Learning. Based on our findings, we follow by describing our proposed FedVSSL approach.
\subsection{Federated Video-SSL System Design}
\label{sec:method_sys}

\begin{figure}[t]
\centering
    \includegraphics[width=\linewidth]{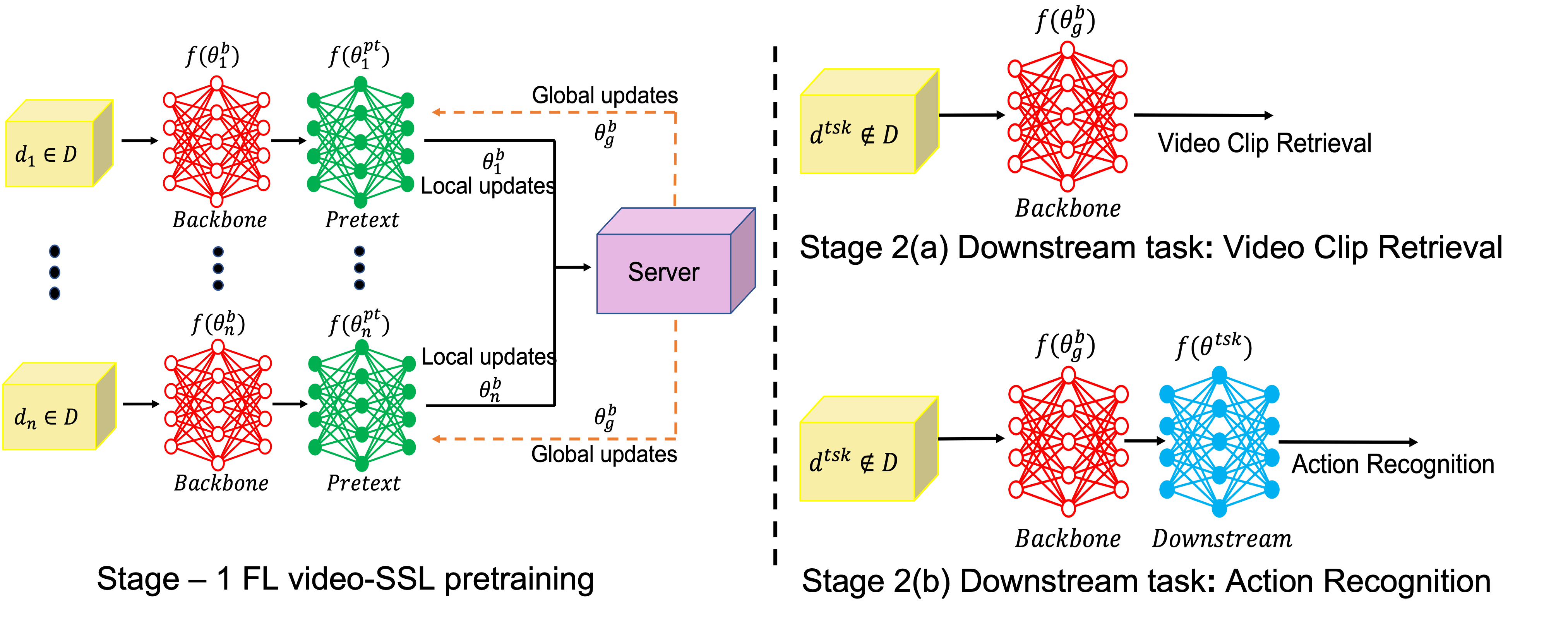}
    \caption{\small System overview: Stage-1 (left) represents the FL video-SSL pretraining. For the downstream tasks; video clip retrieval is seen in Stage-2(a) and action recognition is depicted in Stage-2(b) }
    \label{fig:system_overview}
\end{figure}

We begin with a vanilla FL system that can be integrated with the video-SSL learning. The resulting FL video-SSL system is depicted in Fig.~\ref{fig:system_overview}-Stage 1. We consider having $n$ partitions $\{d_{i}\}_{i=1}^{n}$ of dataset $D$ distributed among $\{c_{i}|c_{i}\in C \}_{i=1}^{n}$  decentralized clients in a Non-IID fashion. Each decentralized client learns intermediate features collaboratively by training the video-SSL approach on their respective local data partition for a few epochs before performing synchronization through the server. The synchronization includes receiving the clients' model parameters, performing model aggregation, and sending back the global model again to clients. More specifically, our training pipeline can be described as follows:

\textbf{1.} Each client $c_i$ holds a set of local parameters $\{\theta^b_{i}, \theta^{p_t}_{i}\}$ for a backbone network $f(\theta^b)$ parameterized by the weights $\theta^b$ followed by a predictor head network $f(\theta^{p_t})$ parameterized by the weights $\theta^{p_t}$. Note that all clients use the same model architecture but the parameter weights $\{\theta^b_{i}, \theta^{p_t}_{i}\}_{1\leq i \leq n}$ can be non-consensus. During each FL round $r$, 
local video-SSL pretraining and synchronization will be conducted intersectively.

\textbf{2.} During the communication/synchronization step, each client would receive global values for the weights $\theta^b_g$ and $\theta^{p_t}_g$ aggregated on the server based on certain FL aggregation strategies.

\textbf{3.} During each round $r$, a random subset of clients, $M$,  is selected to perform video-SSL training. Each participant $m \in M$ optimizes its local model's parameters $\{\theta^b_{m}, \theta^{p_t}_{m}\}$ for $E$ number of local epochs. 

The local training steps in each client completely follows the setup of the video-SSL algorithms, including the pre-training task together with the task-specific loss function design. The optimization is performed with SGD by minimizing the objective $J_{m}^{p_t}(\theta^{b}, \theta^{pt})$:
\begin{equation}
\small J_{m}^{p_t}(\theta^{b}, \theta^{pt}) = \mathbb{E}_{\textbf{x}\sim d_{m}}\mathcal{L}_{m}(\textbf{x}, y_{pt}, \theta^{b},\theta^{pt}),
\label{eq1}
\end{equation}
where $(\textbf{x}, y_{pt})$ is the data with its corresponding pseudo labels for the specific pretext task $p_t$. At the end of the $R$ rounds, the final global model parameters $\theta^{b}_{g}$ for the backbone network can be used to extract spatio-temporal representations from videos for downstream tasks. We then follow the common practice of evaluating the performance of representation learning as done in traditional video-SSL work~\cite{xu2019self,wang2015unsupervised}. The representations learned by the global model parameters $\theta^{b}_{g}$ are evaluated by solving the downstream tasks $tsk$ as is shown in Fig.~\ref{fig:system_overview} - Stage 2. In this work, we consider two downstream tasks: 

\textbf{Action Recognition:} A classification head (fully connected layer) represented by $f(\theta^{tsk})$ is attached following the final layer of $f(\theta^{b}_{g})$, resulting in the model $F_{tsk}$ parameterized by $\{\theta^{b}_{g}, \theta^{tsk} \}$. The $F_{tsk}$ is then fine-tuned in two ways: (1) fine-tuning the whole network $\{\theta^{b}_{g}, \theta^{tsk} \}$, and (2) fine-tuning only the linear classification layer $f(\theta^{tsk}$, i.e., linear probe. 

\textbf{Video Clip Retrieval:} For the video clip retrieval, we determine the video label by majority voting of K-nearest neighbor (KNN) by directly using the parameters learned by $f(\theta^{b}_{g})$ after the FL video-SSL pretraining and followed the protocol as described in \cite{xu2019self}.

The above generic video federated-SSL system allows us to perform a systematic analysis of the video-SSL approaches and its implications in FL  environment against a number of key factors: 

\textbf{Video-SSL in Vanilla FL Settings.} In this scenario, our main objective is to evaluate the performance gap between pretraining video-SSL algorithms in FL and centralized environment. This evaluation will then quantify how different video-SSL methods behave in FL settings with  IID and Non-IID data. 

\textbf{Aggregation Strategies.} The aggregation method plays an important role towards superior performance of the final feature representations learned in FL video-SSL pretraining. We intend to study the effects of different aggregation strategies on the feature representations learned during FL pretraining of video-SSL approaches and hence the downstream tasks. 

\textbf{Loss-Surface and Model Stability of Video-SSL in FL.} These analyses enable us to perform a side-by-side comparison of the key properties of pretraining video-SSL approaches in FL settings against their centralized counterpart. For example, recent work indicates that averaging the weights of the neural networks leads to wider minima \cite{izmailov2018averaging,li2018visualizing} which subsequently provides better generalization. A natural question arises as to whether the aggregation strategies, based on weight averaging, by performing FL video-SSL pretraining could provide wider minima? Moreover, do such wider minima potentially lead to model stability against small-scale perturbations that naturally occur during the communication between edge devices and the server in FL? 

\textbf{Training Efficiency and Communication Cost.} The total time for pretraining video-SSL in FL settings depends on the number of epochs during local training on each client and the number of communication rounds between the clients and the server. It is important to compute the number of communication rounds necessary to achieve a target performance with FL video-SSL training in order to find an optimal balance between the performance of video-SSL and FL system resources.

\subsection{Proposed Method}
Based on the observations of the systematic study of the video-SSL approaches in \textit{cross-device} FL settings, we proposed a novel FL framework, \textit{FedVSSL}, designed specifically for video-SSL in FL  environment. The key characteristics of this framework are as follows: (1) Transceive only the backbone model parameters $\theta^{b}_{g}$ between the server and the clients. (2) FedVSSL aggregation strategy for pretraining video-SSL approaches in FL cross-devicce settings. \\

\noindent \textbf{Transceive only Backbone Weights.} Since each $c_{i}$ client model contains two modules ($f(\theta^b_{c_{i}})$ and $f(\theta^{p_{t}}_{c_{i}})$), we analyse the characteristics of both modules and hypothesize that it is beneficial to only upload and aggregate the parameters $\theta^{b}$ of the backbone model at the server (see Table \ref{Tab-proposed}). The backbone module learns hierarchical deep features from the local data, representing the role of encoder. The classification head is more representative of local data distribution, which learns data-oriented features. We argue that only aggregating the backbone parts could increase generalization of the model, while retaining the classification head and updating it locally could capture more characteristics of non-IID data from clients.
More evidence from Fig. \ref{fig:std_divergence} validates our hypothesis. \\

\noindent \textbf{FedVSSL Aggregation.} The motivation behind proposing FedVSSL aggregation is two folds: First, to integrate the FL aggregation strategies based on weighted averaging under a common framework. Second, to induce the knowledge from the past global models while performing model parameters aggregation.

After local pretraining at round $r$, the server collects the participating clients' locally updated weights ${\theta^{b}_{m}}$, and  the local gradients $g^{(m)}(r)$ from client $m$ at round $r$ is computed as: $g^{(m)}(r) = \theta^{b}_{m}(r) - \theta^{b}_{g}(r - 1)$. The update rule for $\theta^{b}_{g}$ of the global model $F_{g}$ for round $r$ can then be stated as follows:
\begin{equation}
\small \theta^{b}_{g}(r) = \frac{(\sum_{i = 1}^{\beta} \theta_{g}^{b} (r-i)) + \tilde{\theta}^{b}_{g}(r)}{\beta + 1},
\label{eq:SWA}
\end{equation}
where $\tilde{\theta}^{b}_{g}(r) = \theta^{b}_{g}(r - 1) - \eta_s\Delta_r$ and $\eta_s$ is the server learning rate. The averaging is performed over $\beta + 1$ global models. Eq. \ref{eq:SWA} simply represents the stochastic weight averaging (SWA) \cite{izmailov2018averaging} of the global models. As \cite{izmailov2018averaging} reported, simple averaging of the multiple checkpoints of training models obtains better generalisation than conventional training. 
The $\Delta_r$ represents overall local gradients over $m$ clients computed from a weighted combination of different aggregation strategies. Here we choose two aggregation strategies: FedAvg \cite{mcmahan2017communication} and Loss \cite{gao2021end}, which aggregate the client models based on the number of samples or local training loss respectively. Then, $\Delta_r$ can be written as follows:
\begin{equation}
\small \Delta_r = \alpha \Delta_r^{Loss} + (1-\alpha)\Delta_r^{FedAvg}.
\label{eq:weighted_comb}
\end{equation}
The $\alpha$ in Eq. \ref{eq:weighted_comb} controls the amount of the contribution of each aggregation strategy.
Eq. \ref{eq:SWA} is the generalized representation of our FedVSSL aggregation rule. For example, setting $\beta=0$ in Eq. \ref{eq:SWA} reduces the update rule to a weighted combination of FedAvg and Loss. On the other hand, setting $\alpha=0$ or $1$ reduces the update rule to running mean of FedAvg or Loss, respectively. The overall algorithm is summarised in Algo. \ref{al1}.
 
\begin{algorithm}[!t]
            \caption{\small Federated Video Self-supervised Learning (FedVSSL)}
            \label{al1}
            \textbf{Input}: $R, M, N, n_m, \eta_s, \eta_l, \alpha, \beta$ \\
            \textbf{Output}: $\theta^{b}$ \\
            \textbf{Central server does:}
            \begin{algorithmic}[1]
            \For{$r = 1,$...$,R$}
                \State Server randomly sample $M$ clients.
                \For{each $m$ in $M$}
                    \State $\theta_{m}^{b}(r), n_{m}, \mathcal{L}_{m}^{pt}$ =  \textbf{TrainLocally}$(m, \theta^{b}_{g}(r))$
                \EndFor
                
                \State $g^{(m)}(r) = \theta^{b}_{m}(r) - \theta^{b}_{g}(r - 1)$
                \State $\Delta_r^{FedAvg} = \sum_{m = 1}^M \frac{n_m}{\sum_{m=1}^m n_m} g^{(m)}(r)$
                \State $\Delta_r^{Loss} = \sum_{m = 1}^M \frac{\exp{(- \mathcal{L}^{p_t}_{(m)})}}{\sum_{m=1}^M \exp{(- \mathcal{L}^{p_t}_{(m)})}} g^{(m)}(r)$
                \State $\Delta_r = \alpha \Delta_r^{Loss} + (1-\alpha)\Delta_r^{FedAvg}$
                \State Update global model weights $\tilde{\theta}^{b}_{g}(r) \gets \theta^{b}_{g}(r - 1) - \eta_s\Delta_r$.
                \State Compute $\theta^{b}_{g}(r) = \frac{(\sum_{i = 1}^{\beta} \tilde{\theta}_{g}^{b} (r-i)) + \theta^{b}_{g}(r)}{\beta + 1}$.
            \EndFor 
            \end{algorithmic}
            
            \textbf{TrainLocally $(m,\theta^{b}_{g}(r))$:}
            \begin{algorithmic}[1]
            \For{ $k = 1,...,E$}
                \State  $\{\theta^b_m, \theta^{p_t}_{m}\}(k+1) \gets \mathrm{SSL}(\theta^b_{g}(k),\theta^{p_t}_{m}(k), \eta_l)$ based on Eq. \ref{eq1}.
            \EndFor
            \State  Upload $\theta^{b}_{m}, n_{m}, \mathcal{L}_{m}^{pt}$ to the server.
            \end{algorithmic}
\end{algorithm}

%% file: 4_evaluation.tex
\section{Experiments and Results}
In this section, we first describe our experiment setup in Sec. \ref{Dataset_decp} \& Sec. \ref{section:Archiecture and Implementation}). We conduct a systematic analysis of the behavior of video-SSL models in vanilla FL settings in Sec. \ref{sec:vanilla}, followed by a discussion for the results of our proposed FedVSSL in Sec. \ref{sec:propose}.

\subsection{Datasets}
\label{Dataset_decp}
For the pretraining stages of all video-SSL approaches, we utilize kinetics-400 (K400) dataset \cite{kay2017kinetics} with $219$k training samples distributed among $400$ action classes. For downstream task, we utilize the UCF-101 \cite{soomro2012ucf101} (UCF) dataset containing $13,320$ video samples for 101 action classes, and HMDB-51 \cite{Kuehne11} (HMDB) dataset with $7$k video samples distributed among 51 action classes.

\noindent\textbf{Kinectics-400 Non-IID.} Each video sample in K400 comes from a different source, which conforms to the definition of non-IID based on video source-level. To make it more realistic, we generate the non-IID version of K400 based on actual class-labels \cite{mcmahan2017communication}. 
We randomly partition the dataset into 100 shards to mimic the setting of having 100 disjoint clients participating in FL. Each client contains $8$ classes resulting in each client having $2285$ samples on average. Note that there is no overlap of samples between different clients.

\subsection{Architecture and Implementation}
\label{section:Archiecture and Implementation}
\textbf{Video-SSL Approaches.} In this work, we consider three SOTA video-SSL algorithms, all of which propose to solve different pretext tasks for video representation learning. More specifically, VCOP~\cite{xu2019self} learns to determine the permutation order of shuffled clips, Speed~\cite{benaim2020speednet,yao2020video,cho2021self} learns to predict the playback speed of videos, and CtP~\cite{wang2021unsupervised} predicts the positions and sizes of a synthetic image patch in a sequence of video frames. 
For all video-SSL approaches, we use R3D-18 \cite{tran2018closer} architecture as the backbone $f(\theta^{b}$). The architecture choices of prediction heads for different pretext tasks and downstream tasks follow the settings in the original papers. It should be noted that our federated framework is agnostic to different architectures and video-SSL approaches. We develop the FL version of the video-SSL approaches considered in this work on top of Flower \cite{beutel2020flower} federated learning platform by incorporating various video-SSL algorithms developed in MMCV framework \cite{mmcv,wang2021unsupervised}.     
Unless otherwise specified, we keep the settings of the video-SSL approaches as provided by \cite{wang2021unsupervised} during the pretraining tasks and downstream tasks.

\noindent \textbf{FL Pretraining.} We perform the FL pretraining of the video-SSL pretext-task $p_{t}$ using Algo. \ref{al1}. The local pretraining on each client lasts for $E$ epochs per FL round $R$, where we set $E=1$ in our experiments. 
We set the total number of rounds R to $540$ to ensure that each client acquires sufficient participation during FL pretraining.
The selection of the number of $E$ and $R$ is based on our empirical observations. Each round, we randomly select $M=5$ clients from the pool of 100 clients to participate in training and each client trains its local model using SGD optimizer without momentum. We set a constant learning rate of $0.01$ for CtP and Speed and $0.001$ for VCOP. Weight decay is set to $10^{-4}$ and training batch-size is set to 4. On the server side, in addition to our proposed method FedVSSL, we consider three existing aggregation strategies including FedAvg, Loss and FedU.
 
\noindent \textbf{Downstream Tasks.} For the fine-tuning stage of action recognition, we follow the configuration in CtP framework \cite{wang2021unsupervised}. The $F_{d_{tsk}}$ is fine-tuned using the SGD optimizer with an initial learning rate of $0.01$, momentum of $0.9$, and weight decay of $5\times10^{-4}$. The learning rate is decayed by a factor of $0.1$ after $60$ and $120$ epochs, respectively. The batch-size is set $32$, and the fine-tuning stage lasts for 150 epochs. For linear probe, we keep the same settings as for the fine-tuning of the whole network, except that we train only $f(\theta^{d_{tsk}})$ layer of $F_{d_{tsk}}$ for $100$ epochs. The learning rate is decayed by the factor of $0.1$ after $60$ and $80$ epochs. For the video clip retrieval task, we follow the approach described in Sec. \ref{sec:method_sys}.

\setlength{\tabcolsep}{2pt}
\begin{table}[t]
    \caption{\small Action recognition accuracy (Top-1\%) and video clip retrieval accuracy (Top-1\%, Top5\%) on UCF and HMDB for three video-SSL methods. F-T represents fine-tune and L-P stands for linear probe. $\Delta$ represents the difference between centralized and corresponding FL performance.``$+$" and ``$-$" show \% improvement and degradation respectively. $*$ means the results are reproduced using the implementation in \cite{wang2021unsupervised}}
    \label{Tab-action_recog_comp}
    \centering
    \scalebox{0.85}{
    \begin{tabular}{l cccc cccc}
    \toprule
     & \multicolumn{4}{c}{\textbf{Action Recognition}} & \multicolumn{4}{c}{\textbf{Video Clip Retrieval}}\\
   \cmidrule(r){2-5} \cmidrule(r){6-9}
      & \multicolumn{2}{c}{UCF} & \multicolumn{2}{c}{HMDB} & \multicolumn{2}{c}{UCF} &  \multicolumn{2}{c}{HMDB}\\
      \cmidrule(r){2-3} \cmidrule(r){4-5} \cmidrule(r){6-7} \cmidrule(r){8-9}
     \textbf{SSL Method} & F-T & L-P & F-T & L-P & R@1 & R@5 & R@1 & R@5\\ 
     \cmidrule(r){1-9}
    VCOP* & 71.29 & 24.93 & 38.56 & 13.53 &  15.52 & 28.26 & 8.11 & 22.22  \\
     VCOP(Fed)& 69.26 &  20.00 & 33.27 & 12.22 &  13.72 & 24.85 & 6.41 & 19.94  \\
    $\Delta$ & (-2.03) & (-4.93) &(-5.29) &(-1.31) &(-1.8) & (-3.41) & (-1.7) & (-2.28) \\
    \cmidrule(r){1-9}
    Speed*  & 81.15 & 29.32 & 47.58 & 14.90  & 16.84 & 36.58  & 6.93 & 21.05 \\
    Speed(Fed)  & 73.16 & 35.63& 38.43 & 21.57 & 21.97 & 41.61 & 10.98 & 28.30 \\
    $\Delta$ & (-7.99) &  (+6.31) & (-9.05) & (+6.67) & (+4.05) & (+3.94) & (+3.92) & (+7.25) \\
    \cmidrule(r){1-9}
    CtP*  & \textbf{86.20} & \textbf{48.14} & \textbf{57.00} & \textbf{30.65} & 29.0 & 47.30 & 11.80 & 30.10  \\
    CtP(Fed) & 81.95 & 46.13& 49.15 & 28.63 & \textbf{29.29} & \textbf{48.90} & \textbf{13.66} & \textbf{32.42} \\
    $\Delta$ & (-4.25) & (-2.01) & (-7.85) & (-2.02) & (+0.29) & (+1.6) & (+1.86) & (+2.32) \\
    \bottomrule
   \end{tabular}
   }
\end{table}

\subsection{Video-SSL in Vanilla FL Settings}
\label{sec:vanilla}
Here we investigate the performance of video-SSL in vanilla FL settings against the key factors listed in Sec.\ref{sec:method_sys}. 

\noindent \textbf{Centralized vs. Federated Video-SSL.} In this experiment, we draw a first investigation for the performance of video-SSL approaches using FedAvg in centralized and vanilla FL settings with Non-IID video data. We report this comparison in terms of three downstream tasks, i.e, fine-tuning, linear probe, video clip retrieval, on UCF and HMDB datasets (Table \ref{Tab-action_recog_comp}).
First, CtP obtains the best performance for all tasks in both centralized and FL settings.
Second, when the trained network is fine-tuned for the action recognition task, the centralized video-SSL approaches perform better compared to their corresponding FL counterparts. However, the degradation in the performance is not drastic. We conjecture that it is caused by the smoother and flatter manifold in the models in FL settings, which would be more challenging to fine-tune.
Third, the linear-probe results for action recognition show mixed results due to the fact that only the classification head participates in the training.
Finally, the video clip retrieval results are more competitive with the FL version of Speed and CtP, which achieve better performance than their centralized counterparts.

Given the fact that the video-SSL benefits from the large-scale datasets and the performance degradation of the FL version of the video-SSL is acceptable (even in the Non-IID case with FedAvg), it makes the FL a natural choice for video-SSL approaches. 

\begin{table}[t]
    \caption{\small Action recognition and video clip retrieval accuracies (\%) on UCF and HMDB for the federated CtP models pretrained with one local epoch per round. C represents the number of clients, and Cpc stands for the number of classes per client} 
    \label{Tab-ablation_action}
    \centering
    \scalebox{0.85}{
    \begin{tabular}{lcccc cccc}
        \toprule
         &  & & \multicolumn{2}{c}{\textbf{Fine-tuning}} & \multicolumn{4}{c}{\textbf{Retrieval}}  \\
        \cmidrule(r){4-5} \cmidrule(r){6-9}
        & & & UCF & HMDB & \multicolumn{2}{c}{UCF} & \multicolumn{2}{c}{HMDB}\\
        \cmidrule(r){4-4} \cmidrule(r){5-5} \cmidrule(r){6-7} \cmidrule(r){8-9}
       \textbf{Method} & \textbf{C/Cpc} & \textbf{Data}  & Top-1 & Top-1 & R@1 & R@5 & R@1 & R@5 \\
        \cmidrule(r){1-9}
        \multirow{4}{4em}{CtP(Fed.)} & 100/- & IID & 81.92 & 48.49 & 29.42 & 47.90 & 13.80 & 34.56\\ 
& 100/8 & Non-IID & 81.95 & 49.15 & 29.29 & 48.90 &  13.66 & 32.42\\
& 100/4 & Non-IID & 81.15 & 47.78 & 29.18 & 48.37 & 14.70 & 32.94 \\
        \hline
        CtP(Cent.) & - & IID & 86.2 & 57.00 & 29.00 & 47.30 & 11.80 & 30.10 \\
        \bottomrule
    \end{tabular}
    }
\end{table}

\noindent \textbf{Performance with IID vs. Non-IID Data.} Conventional FL methods are designed to solve a supervised/semi-supervised learning task within an IID/Non-IID data distribution based on the actual class-labels. 
To understand the impact of data distribution on federated video-SSL training, we simulate IID and Non-IID settings based on class-labels and compare the performance of downstream task in Table \ref{Tab-ablation_action}. 
The Non-IID versions of K-400 are generated with two variations of the distribution of the samples among 100 clients, with 4 and 8 classes per client respectively.
We report the fine-tuning and video clip retrieval accuracy on UCF and HMDB, by pretraining CtP video-SSL approach on all settings.

One can see from Table \ref{Tab-ablation_action} that pretraining the CtP video-SSL approach using standard FedAvg with IID and different degrees of Non-IID levels achieve comparatively similar performance on the fine-tuning and video clip retrieval task.
This could be explained by the generation process of IID/Non-IID data based the actual class-labels. The video-SSL methods learn representations by generating pseudo labels based on the pretext task it solves, which is independent of actual class-labels. Hence, the IID/Non-IID data distribution has a slight impact to the SSL model training.
In addition, we find that there exists a degradation in the fine-tuning performance of federated CtP video-SSL compared to its centralized counterpart. Interestingly, the FL version of CtP video-SSL approach gives better video clip retrieval performance when compared to its centralized counterpart with both IID and different degrees of Non-IID levels.

\noindent \textbf{Performance of Aggregation Strategies.} The aggregation method often plays an important role towards superior performance of models trained in FL environments. 
In this experiment, we investigate the impact of FL aggregation strategies on the pretraining by evaluating the final performance of video-SSL approaches in Non-IID FL settings. In Table \ref{Tab-action_finetune_adam}, we show the performance of CtP video-SSL approach on UCF and HMDB against a range of aggregation strategies that include FedAvg \cite{mcmahan2017communication}, Loss \cite{gao2021end}, and FedU \cite{zhuang2021collaborative}. 
It can be observed that FedAvg performs better on HMDB and obtains similar performance with Loss method on UCF. Additionally, except for retrieval accuracy on HMDB, FedU outperforms others on both video clip retrieval and fine-tuning mainly due to its dynamic aggregation mechanism. 

\setlength{\tabcolsep}{1.7pt}
\begin{table}[t]
\begin{center}
    \caption{\small Video clip retrieval accuracies (\%) and fine-tuning accuracies (\%) on UCF and HMDB for CtP video-SSL approach using various aggregation strategies. The SSL pretraining is performed on K400 (Non-IID).
    Cent$^\dagger$ represents the centralized training for $27$ epochs which equals to $540$ rounds in our FL setting } 
    \label{Tab-action_finetune_adam}
    \scalebox{0.85}{
    \begin{tabular}{l cccc cc}
    \toprule
      & \multicolumn{4}{c}{\textbf{Retrieval}} & \multicolumn{2}{c}{\textbf{Fine-tuning}} \\
     \cmidrule(r){2-5} \cmidrule(r){6-7}
      & \multicolumn{2}{c}{UCF} & \multicolumn{2}{c}{HMDB} & UCF & HMDB \\
   \cmidrule(r){2-3} \cmidrule(r){4-5} \cmidrule(r){6-6} \cmidrule(r){7-7} 
    \textbf{Method} &  R@1 & R@5 & R@1 & R@5 & Top-1 & Top-1 \\
     \cmidrule(r){1-7}
      Centralized & 29.00 & 47.3 & 11.80 & 30.1 & 86.20 & 57.00 \\
      Centralized$^\dagger$ & 27.65 & 47.67 & 12.81 & 31.05 & 83.64 & 53.73 \\
       \cmidrule(r){1-7}
      FedAvg (Baseline) & 32.62 & 50.41 & \textbf{16.54} & 35.29 & 79.91 & 52.88 \\
      Loss-based &  32.54 & 50.01 & 14.44 &  34.97 & 79.43 & 50.63  \\
      FedU &   \textbf{34.07} & \textbf{52.29} & 14.90 & \textbf{36.67} & \textbf{80.17} & \textbf{53.73} \\
     \bottomrule
    \end{tabular}
    }
\end{center}
\end{table}

\begin{figure}[t]
    \centering
    \begin{subfigure}{0.3\textwidth}
    \centering
    \includegraphics[width=\linewidth]{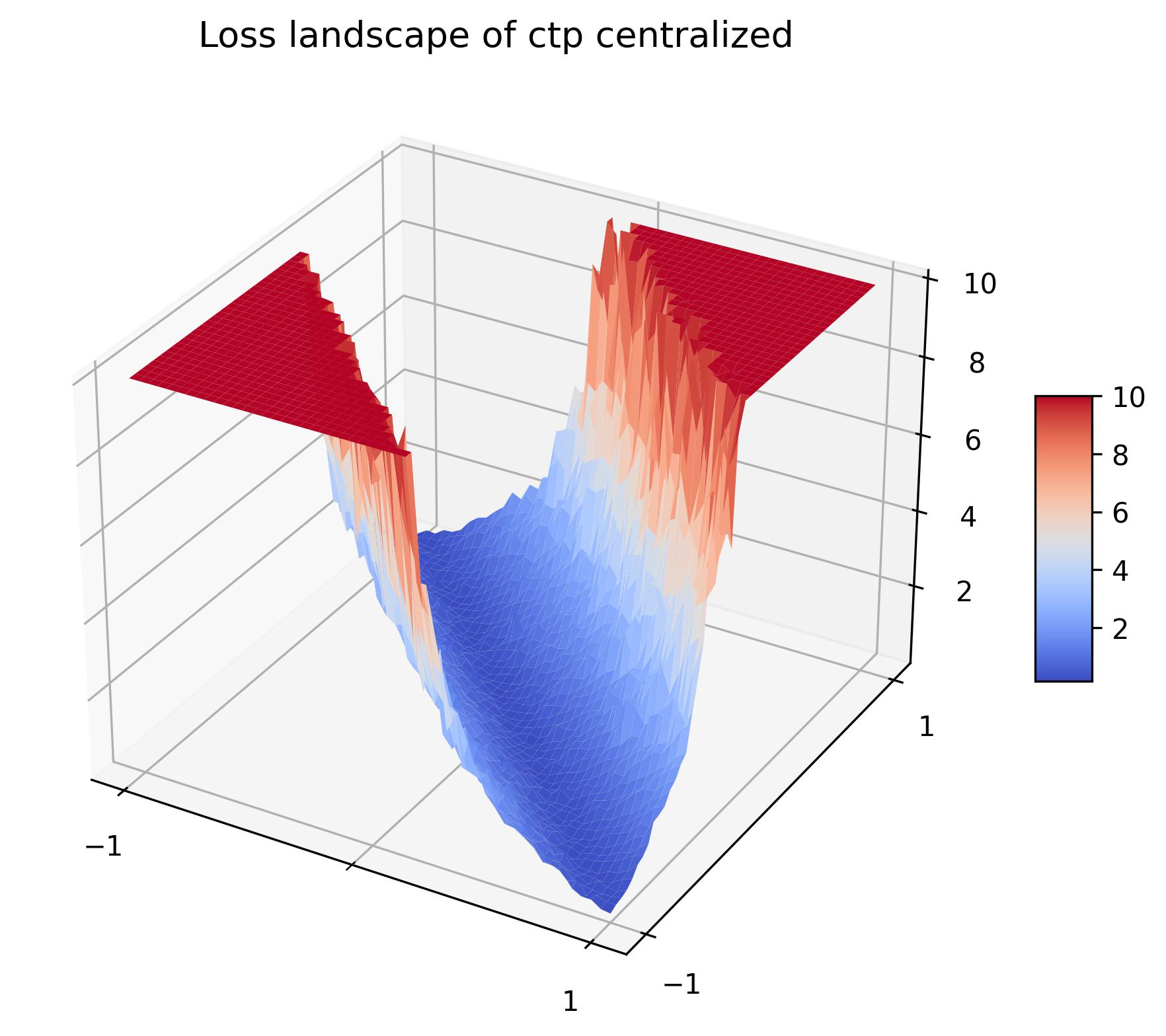}
    \end{subfigure}%
    \begin{subfigure}{0.3\textwidth}
    \centering
    \includegraphics[width=\linewidth]{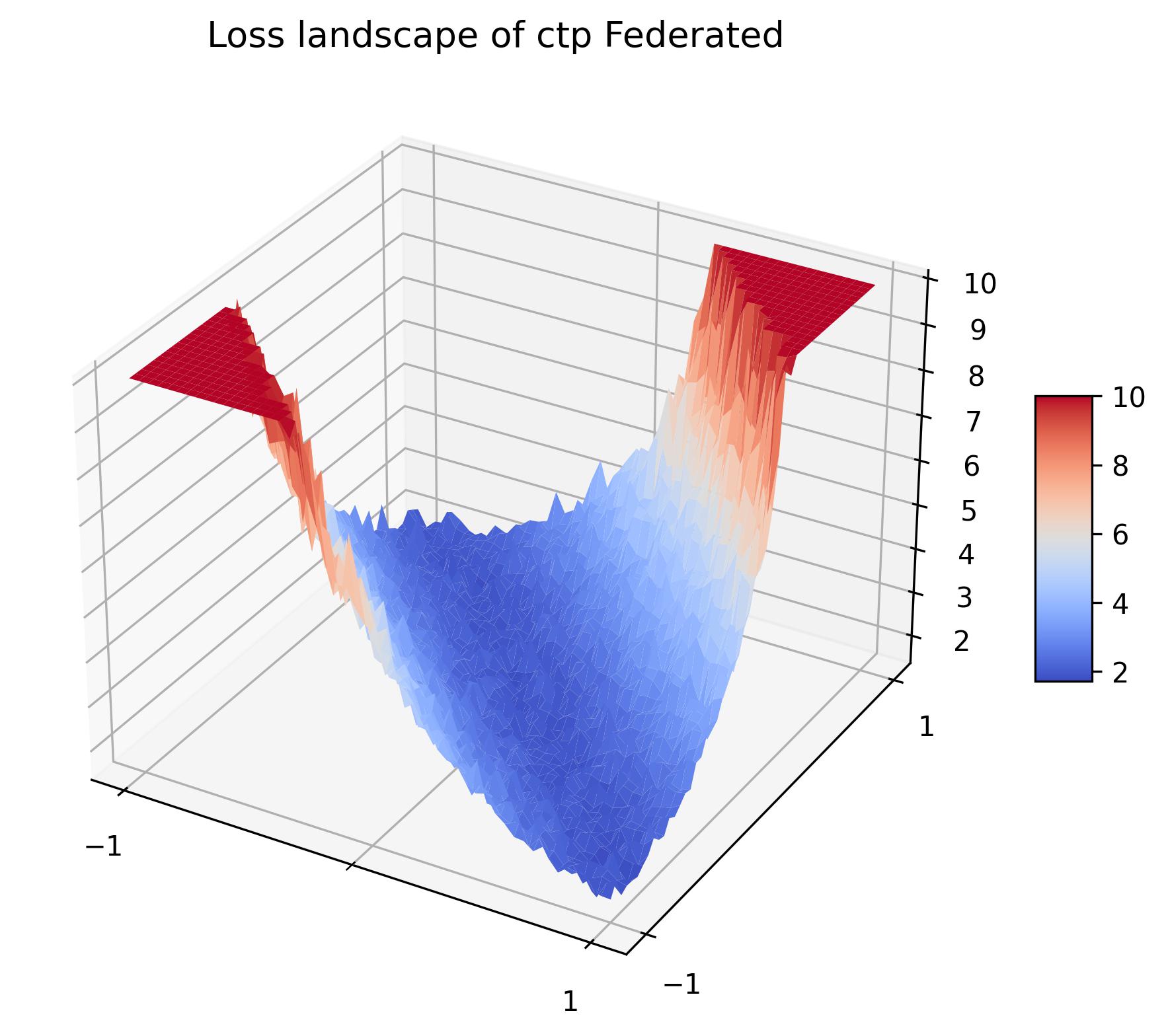}
    \end{subfigure}%
    
    \caption{\small Loss landscape of the final $f(\theta^{b})$ pretrained in a centralized(left) and FL with FedAvg(right) settings}
    \label{fig:cent_loss_surf}
\end{figure}

\noindent \textbf{Loss Surface and Model Stability of Video-SSL in FL.}
To understand why the federated video-SSL models gain higher retrieval accuracy than centralized models, we further analyze the loss landscape around the pretrained model both in centralized and FL settings. To compute this, we utilize the filter normalization method as proposed in \cite{li2018visualizing}. The results are shown in Fig. \ref{fig:cent_loss_surf} for CtP video-SSL approach. We find that the loss landscape of model pretrained with FedAvg is flatter than the model pretrianed with the centralized video-SSL. 
The width of the optima is critically related to generalization, which enables the model to converge in a point centered in this region \cite{izmailov2018averaging,keskar2016large,hochreiter1997flat}. This often leads to slightly worse training loss but substantially better test accuracy.

\begin{figure}[pt]
    \centering
    
     \begin{subfigure}{0.5\textwidth}
    \centering
    \includegraphics[width=\linewidth]{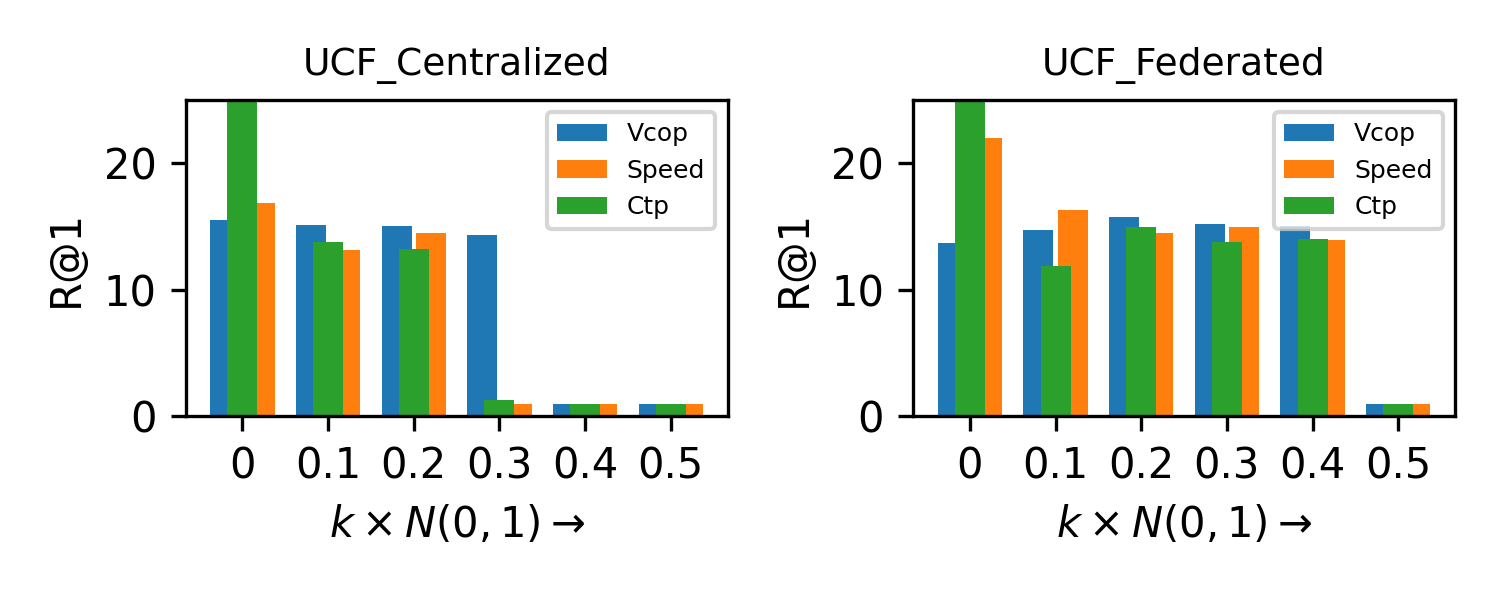}%
    \\
    \end{subfigure}%
    \begin{subfigure}{0.5\textwidth}
    \centering
    \includegraphics[width=\linewidth]{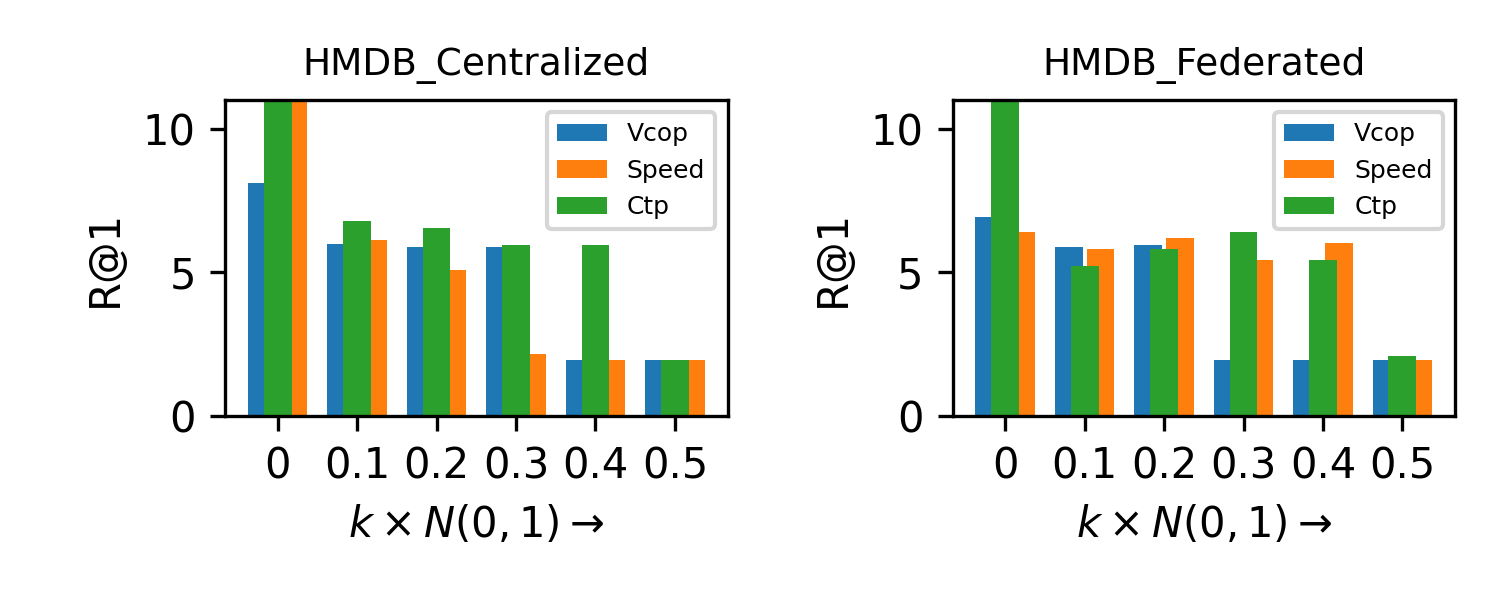}
    \end{subfigure}
    \caption{\small Top1\% action retrieval accuracy on UCF and HMDB by adding perturbation to all the weights of the pretrained $f(\theta^{b})$ network. The perturbation is sampled from a normal distribution $N(0,1)$ and multiplied by a factor $k$}
    \label{fig:perturb_fed_vs_cent}
\end{figure}

We then explore whether such wider optima could increase model stability against small perturbations. To achieve this, we perturb the weights of the backbone pretrained with centralized and federated video-SSL approach. The perturbations are sampled from a uniform normal distribution with zero mean and unit variance, i.e., $\mathcal{N}(0,1)$. We start with the perturbation level of $0$ and incrementally increase the level of perturbation by a factor of $0.1$, i.e., ($\{\mathcal{N}(0,1)\times x |0\leq x \leq0.5\}$). 
The results are shown in Fig. \ref{fig:perturb_fed_vs_cent} for the top-1\% video clip retrieval accuracy for the centralized video-SSL approach and its corresponding FL counterparts, on UCF and HMDB datasets. 
One can see from that as the level of perturbation is increased from 0 to 0.1 the top-1\% accuracy drops significantly for both centralized CtP video-SSL and its FL counterparts. However, as the level for perturbation is further increased we find that the FL versions of the video-SSL approach show good stability compared to its centralized counterpart on both datasets. Overall, the federated training boosts the generalization and stability of video-SSL models. 

\begin{figure}[pt]
    \centering
     \begin{subfigure}{0.5\textwidth}
    \centering
    \includegraphics[width=0.9\linewidth]{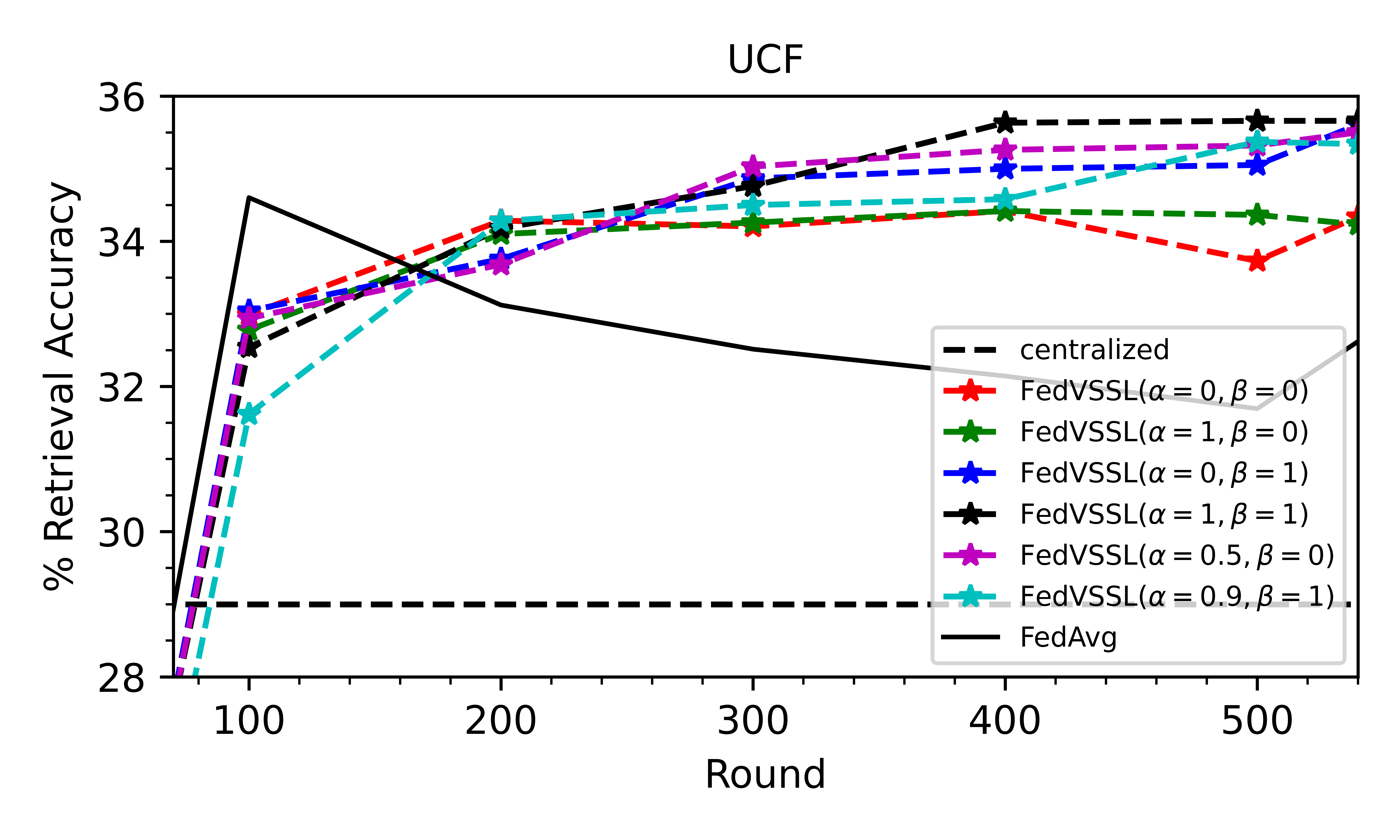}%
    \\
    \end{subfigure}%
    \begin{subfigure}{0.5\textwidth}
    \centering
    \includegraphics[width=0.9\linewidth]{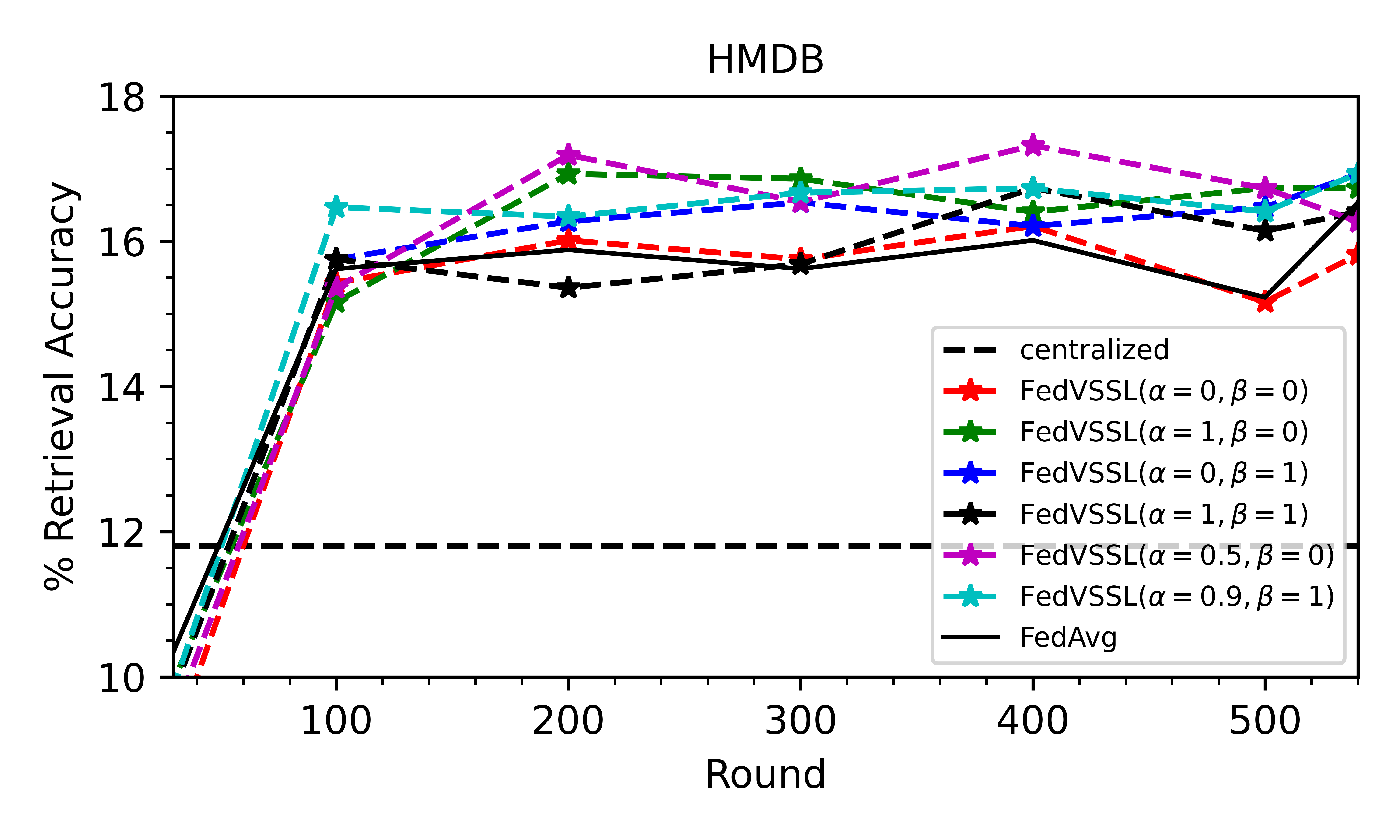}
    \end{subfigure}
    \caption{\small Top $1\%$ retrieval accuracy ($\%$) w.r.t communication rounds for our proposed FL methods on UCF(left) and HMDB (right).  Our proposed methods require less than 200 rounds to reach the centralized target accuracy on both datasets}
    \label{fig:comm_vs_no_rounds}
\end{figure}

\noindent \textbf{Training  Efficiency and Communication Cost.} In this experiment, we analyze the computational efficiency of pretraining CtP approach in both FL and centralized settings. We find that our FL pretraining of CtP for $540$ rounds, with $100$ clients and $5\%$ client sampling rate (in ideal scenarios), is equivalent to $27$ epochs ($30\%$) of centralized pretraining. This results in $62.5\%$ of GPU time saving compared to the centralized pretraining of the CtP that lasts for $90$ epochs. Additionally, compared with the performance of centralized pretraining of CtP for $27$ epochs, our federated CtP model achieves significant boost in video clip retrieval performance while providing competitive performance on fine-tuning as shown in Table \ref{Tab-action_finetune_adam}. 
We further show the number of communication rounds required by federated video-SSL models to achieve the centralized target video clip retrieval accuracy. 
One can see from Fig. \ref{fig:comm_vs_no_rounds} that the FL model trained with FedAvg requires less than $100$ rounds to reach the centralized target accuracy on UCF and HMDB datasets. Indeed, our proposed FedVSSL shows even better better performance and convergence behaviour.

\subsection{Results on FedVSSL}
\label{sec:propose}

In this section, we investigate the performance of our proposed FVSSL method, conduct ablation studies by varying $\alpha$ and $\beta$ in Eq. \ref{eq:SWA} and Eq. \ref{eq:weighted_comb}, and report the results on UCF and HMDB in Table \ref{Tab-proposed}. One can see that the performance of all proposed methods are competitive, which demonstrates that the learned representations are qualified for the downstream applications.

\begin{figure}[h]
    \centering
    \includegraphics[width=0.7\linewidth]{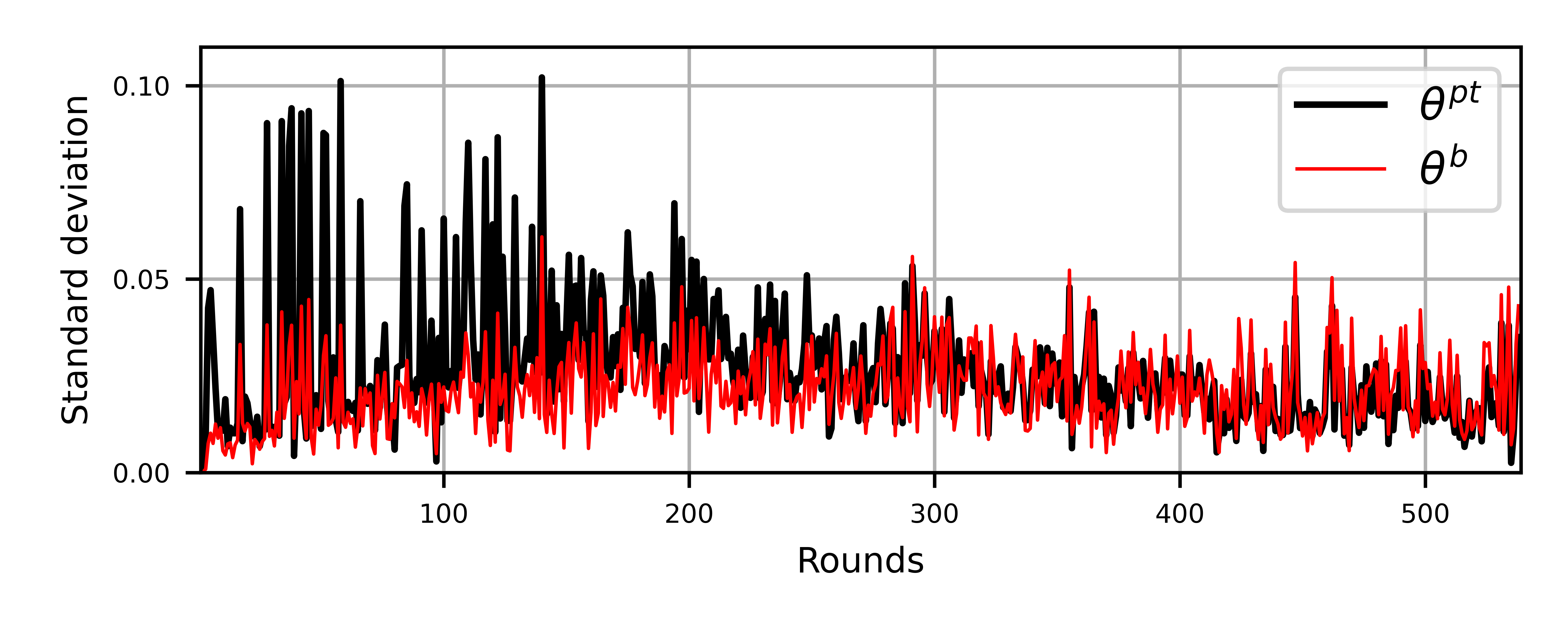}
    \caption{\small Standard deviations of the $L_{2}$ difference between the global model weights and the locally trained model weights at each round of FL video-SSL pretraining with FedAvg. Both backbone $\theta^b$ and prediction head $\theta^{p_t}$ are aggregated on the server}
    \label{fig:std_divergence}
\end{figure}

\begin{table}[h]
\begin{center}
    \caption{\small Video clip retrieval accuracies (\%) and fine-tuning accuracies (\%) on UCF101 and HMDB51 for CtP video-SSL approach using our proposed FedVSSL methods. The SSL pretraining is performed on K400 (Non-IID)} 
    \label{Tab-proposed}
    \scalebox{0.85}{
    \begin{tabular}{l cccc cc cc}
    \toprule
      & \multicolumn{4}{c}{\textbf{Retrieval}} & \multicolumn{2}{c}{\textbf{Fine-tuning}} & \multicolumn{2}{c}{\textbf{Linear-probe}} \\
      \cmidrule(r){2-5} \cmidrule(r){6-7} \cmidrule(r){8-9}
      & \multicolumn{2}{c}{UCF} & \multicolumn{2}{c}{HMDB} & UCF & HMDB & UCF & HMDB \\
    \cmidrule(r){2-3} \cmidrule(r){4-5} \cmidrule(r){6-6} \cmidrule(r){7-7} \cmidrule(r){8-8} \cmidrule(r){9-9}
    \textbf{Method} &  R@1 & R@5 & R@1 & R@5 & Top-1 & Top-1 & Top-1 & Top-1 \\
     \cmidrule(r){1-9}
      Centralized & 29.00 & 47.3 & 11.80 & 30.1 & 86.20 & 57.00 & 48.14 & 30.65 \\
     \cmidrule(r){1-9}
     FedAvg (Baseline) & 32.62 & 50.41 & 16.54 & 35.29 & 79.91 & 52.88 & 45.31 & 31.44 \\
     FedVSSL($\alpha=0,\beta=0$) & 34.34 & 51.71 & 15.82 & 36.01 & 79.91 & 52.94 & 47.95 & 31.12 \\
     FedVSSL($\alpha=1,\beta=0$)& 34.23 & 52.21 & 16.73& \textbf{38.30} & 79.14 & 51.11 & 47.90 & 29.48 \\
     FedVSSL($\alpha=0,\beta=1$) &  35.61 & 52.18 & \textbf{16.93} & 37.78 & 79.43 & 51.90 & 47.66 & 30.00 \\
     FedVSSL($\alpha=1,\beta=1$)& \textbf{35.66} & 52.34 & 16.41 & 36.93 & 78.99 & 51.18 & 48.93 & 31.44 \\
     FedVSSL($\alpha=0.9,\beta=0$)&  35.50 & \textbf{54.27} & 16.27& 37.25 & \textbf{80.62} & \textbf{53.14} & \textbf{50.36} & \textbf{32.68} \\
     FedVSSL($\alpha=0.9,\beta=1$)&  35.34 & 52.34 & \textbf{16.93}& 37.39 & 79.41 & 51.50 & 50.30 & 32.42 \\
     \bottomrule
    \end{tabular}
    }
\end{center}
\end{table}

First, all models trained with FedVSSL provide superior retrieval performance.
Concretely, (FedVSSL with $\alpha=0.9, \beta=0$) we improve the video clip retrieval performance over FedAvg baseline by $2.88\%$ (top$1\%$) and $3.86\%$ (top$5\%$) on UCF, and $1.96\%$ (top$5\%$) on HMDB. We obtained $0.71\%$ and $0.26\%$ improvement (top1\%) in the category of fine-tuning  on UCF and HMDB, respectively compared to FedAvg baseline. In the category of Linear-probe, we obtained $5.05\%$ and $1.24\%$ improvement (top1\%) on UCF and HMDB, respectively compared to the FedAvg baseline.  
Second, the component of SWA ($\beta=1$)in FedVSSL has a distinct benefit on the improvement of retrieval performance. 
Third, the top-1\% retrieval accuracy of FedVSSL($\alpha=0,\beta=0$) outperforms the FedAvg baseline by $1.72\%$ on UCF dataset, which highlights the benefits of only updating backbone.

One can see from Fig. \ref{fig:std_divergence} that the standard deviation of the $L_{2}$ distance for the backbone model parameter is more consistent throughout the FL pretraining compared to that of prediction model parameters. This indicates that the backbone weights cause less divergence in FL pretraining of video-SSL which may provide significant efficiency and performance boost in the FL scenarios with stringent communication budget.

%% file: 5_conclusions.tex
\section{Conclusions}

In this paper, we presented the first systematic study on video-SSL for FL. Our key findings  include (1) the importance of aggregating just the backbone network and that (2) non-IID definition based on class-labels bears no impact on pretext training as they are not used by that task. Based on these findings, we proposed FedVSSL, an aggregation strategy tailored to video. FedVSSL was able to outperform the centralized SOTA for the downstream retrieval task by $6.66$\% on UCF-101 dataset and by $5.13$\% on HMDB-51 using non-contrastive methods. We hope this work will enable future research towards further combining FL and SSL for  video representation learning. These are complementary technologies that can together harvest rich visual information from edge devices while still preserving user privacy.

%% file: 0_eccv2022submission.bbl
\begin{thebibliography}{10}
\providecommand{\url}[1]{\texttt{#1}}
\providecommand{\urlprefix}{URL }
\providecommand{\doi}[1]{https://doi.org/#1}

\bibitem{aytar2016soundnet}
Aytar, Y., Vondrick, C., Torralba, A.: Soundnet: Learning sound representations
  from unlabeled video. Advances in neural information processing systems
  \textbf{29} (2016)

\bibitem{benaim2020speednet}
Benaim, S., Ephrat, A., Lang, O., Mosseri, I., Freeman, W.T., Rubinstein, M.,
  Irani, M., Dekel, T.: Speednet: Learning the speediness in videos. In:
  Proceedings of the IEEE/CVF Conference on Computer Vision and Pattern
  Recognition. pp. 9922--9931 (2020)

\bibitem{beutel2020flower}
Beutel, D.J., Topal, T., Mathur, A., Qiu, X., Parcollet, T., Lane, N.D.:
  Flower: A friendly federated learning research framework. arXiv preprint
  arXiv:2007.14390  (2020)

\bibitem{chen2020simple}
Chen, T., Kornblith, S., Norouzi, M., Hinton, G.: A simple framework for
  contrastive learning of visual representations. In: International conference
  on machine learning. pp. 1597--1607. PMLR (2020)

\bibitem{cho2021self}
Cho, H., Kim, T., Chang, H.J., Hwang, W.: Self-supervised visual learning by
  variable playback speeds prediction of a video. IEEE Access  (2021)

\bibitem{mmcv}
Contributors, M.: {MMCV: OpenMMLab} computer vision foundation.
  \url{https://github.com/open-mmlab/mmcv} (2018)

\bibitem{doersch2015unsupervised}
Doersch, C., Gupta, A., Efros, A.A.: Unsupervised visual representation
  learning by context prediction. In: Proceedings of the IEEE international
  conference on computer vision. pp. 1422--1430 (2015)

\bibitem{feichtenhofer2021large}
Feichtenhofer, C., Fan, H., Xiong, B., Girshick, R., He, K.: A large-scale
  study on unsupervised spatiotemporal representation learning. In: Proceedings
  of the IEEE/CVF Conference on Computer Vision and Pattern Recognition. pp.
  3299--3309 (2021)

\bibitem{gao2021end}
Gao, Y., Parcollet, T., Zaiem, S., Fernandez-Marques, J., de~Gusmao, P.P.,
  Beutel, D.J., Lane, N.D.: End-to-end speech recognition from federated
  acoustic models. arXiv preprint arXiv:2104.14297  (2021)

\bibitem{goyal2022vision}
Goyal, P., Duval, Q., Seessel, I., Caron, M., Singh, M., Misra, I., Sagun, L.,
  Joulin, A., Bojanowski, P.: Vision models are more robust and fair when
  pretrained on uncurated images without supervision. arXiv preprint
  arXiv:2202.08360  (2022)

\bibitem{han2019video}
Han, T., Xie, W., Zisserman, A.: Video representation learning by dense
  predictive coding. In: Proceedings of the IEEE/CVF International Conference
  on Computer Vision Workshops. pp.~0--0 (2019)

\bibitem{han2020self}
Han, T., Xie, W., Zisserman, A.: Self-supervised co-training for video
  representation learning. Advances in Neural Information Processing Systems
  \textbf{33},  5679--5690 (2020)

\bibitem{he2020momentum}
He, K., Fan, H., Wu, Y., Xie, S., Girshick, R.: Momentum contrast for
  unsupervised visual representation learning. In: Proceedings of the IEEE/CVF
  Conference on Computer Vision and Pattern Recognition. pp. 9729--9738 (2020)

\bibitem{hochreiter1997flat}
Hochreiter, S., Schmidhuber, J.: Flat minima. Neural computation
  \textbf{9}(1),  1--42 (1997)

\bibitem{hu2022dynamic}
Hu, Z., Xie, H., Yu, L., Gao, X., Shang, Z., Zhang, Y.: Dynamic-aware federated
  learning for face forgery video detection. ACM Transactions on Intelligent
  Systems and Technology (TIST)  (2022)

\bibitem{izmailov2018averaging}
Izmailov, P., Podoprikhin, D., Garipov, T., Vetrov, D., Wilson, A.G.: Averaging
  weights leads to wider optima and better generalization. In: 34th Conference
  on Uncertainty in Artificial Intelligence 2018, UAI 2018. pp. 876--885.
  Association For Uncertainty in Artificial Intelligence (AUAI) (2018)

\bibitem{jain2021biometrics}
Jain, A.K., Deb, D., Engelsma, J.J.: Biometrics: Trust, but verify. arXiv
  preprint arXiv:2105.06625  (2021)

\bibitem{jenni2020video}
Jenni, S., Meishvili, G., Favaro, P.: Video representation learning by
  recognizing temporal transformations. In: Computer Vision--ECCV 2020: 16th
  European Conference, Glasgow, UK, August 23--28, 2020, Proceedings, Part
  XXVIII 16. pp. 425--442. Springer (2020)

\bibitem{jing2018self}
Jing, L., Yang, X., Liu, J., Tian, Y.: Self-supervised spatiotemporal feature
  learning via video rotation prediction. arXiv preprint arXiv:1811.11387
  (2018)

\bibitem{kairouz2019advances}
Kairouz, P., McMahan, H.B., Avent, B., Bellet, A., Bennis, M., Bhagoji, A.N.,
  Bonawitz, K., Charles, Z., Cormode, G., Cummings, R., et~al.: Advances and
  open problems in federated learning. arXiv preprint arXiv:1912.04977  (2019)

\bibitem{kay2017kinetics}
Kay, W., Carreira, J., Simonyan, K., Zhang, B., Hillier, C., Vijayanarasimhan,
  S., Viola, F., Green, T., Back, T., Natsev, P., et~al.: The kinetics human
  action video dataset. arXiv preprint arXiv:1705.06950  (2017)

\bibitem{keskar2016large}
Keskar, N.S., Mudigere, D., Nocedal, J., Smelyanskiy, M., Tang, P.T.P.: On
  large-batch training for deep learning: Generalization gap and sharp minima.
  arXiv preprint arXiv:1609.04836  (2016)

\bibitem{kolesnikov2019revisiting}
Kolesnikov, A., Zhai, X., Beyer, L.: Revisiting self-supervised visual
  representation learning. In: Proceedings of the IEEE/CVF conference on
  computer vision and pattern recognition. pp. 1920--1929 (2019)

\bibitem{krizhevsky2009learning}
Krizhevsky, A.: Learning multiple layers of features from tiny images  (2009)

\bibitem{Kuehne11}
Kuehne, H., Jhuang, H., Garrote, E., Poggio, T., Serre, T.: {HMDB}: a large
  video database for human motion recognition. In: Proceedings of the
  International Conference on Computer Vision (ICCV) (2011)

\bibitem{lee2017unsupervised}
Lee, H.Y., Huang, J.B., Singh, M., Yang, M.H.: Unsupervised representation
  learning by sorting sequences. In: Proceedings of the IEEE International
  Conference on Computer Vision. pp. 667--676 (2017)

\bibitem{li2018visualizing}
Li, H., Xu, Z., Taylor, G., Studer, C., Goldstein, T.: Visualizing the loss
  landscape of neural nets. In: Proceedings of the 32nd International
  Conference on Neural Information Processing Systems. pp. 6391--6401 (2018)

\bibitem{li2020learning}
Li, T., Wang, L.: Learning spatiotemporal features via video and text pair
  discrimination. arXiv preprint arXiv:2001.05691  (2020)

\bibitem{mcmahan2017communication}
McMahan, B., Moore, E., Ramage, D., Hampson, S., y~Arcas, B.A.:
  Communication-efficient learning of deep networks from decentralized data.
  In: Artificial intelligence and statistics. pp. 1273--1282. PMLR (2017)

\bibitem{misra2016shuffle}
Misra, I., Zitnick, C.L., Hebert, M.: Shuffle and learn: unsupervised learning
  using temporal order verification. In: European Conference on Computer
  Vision. pp. 527--544. Springer (2016)

\bibitem{park2020sinet}
Park, H., Sjosund, L., Yoo, Y., Monet, N., Bang, J., Kwak, N.: Sinet: Extreme
  lightweight portrait segmentation networks with spatial squeeze module and
  information blocking decoder. In: Proceedings of the IEEE/CVF Winter
  Conference on Applications of Computer Vision. pp. 2066--2074 (2020)

\bibitem{piergiovanni2020evolving}
Piergiovanni, A., Angelova, A., Ryoo, M.S.: Evolving losses for unsupervised
  video representation learning. In: Proceedings of the IEEE/CVF Conference on
  Computer Vision and Pattern Recognition. pp. 133--142 (2020)

\bibitem{reddi2020adaptive}
Reddi, S.J., Charles, Z., Zaheer, M., Garrett, Z., Rush, K., Kone{\v{c}}n{\`y},
  J., Kumar, S., McMahan, H.B.: Adaptive federated optimization. In:
  International Conference on Learning Representations (2020)

\bibitem{romijnders2021representation}
Romijnders, R., Mahendran, A., Tschannen, M., Djolonga, J., Ritter, M.,
  Houlsby, N., Lucic, M.: Representation learning from videos in-the-wild: An
  object-centric approach. In: Proceedings of the IEEE/CVF Winter Conference on
  Applications of Computer Vision. pp. 177--187 (2021)

\bibitem{russakovsky2015imagenet}
Russakovsky, O., Deng, J., Su, H., Krause, J., Satheesh, S., Ma, S., Huang, Z.,
  Karpathy, A., Khosla, A., Bernstein, M., et~al.: Imagenet large scale visual
  recognition challenge. International journal of computer vision
  \textbf{115}(3),  211--252 (2015)

\bibitem{soomro2012ucf101}
Soomro, K., Zamir, A.R., Shah, M.: Ucf101: A dataset of 101 human actions
  classes from videos in the wild  (2012)

\bibitem{tran2018closer}
Tran, D., Wang, H., Torresani, L., Ray, J., LeCun, Y., Paluri, M.: A closer
  look at spatiotemporal convolutions for action recognition. In: Proceedings
  of the IEEE conference on Computer Vision and Pattern Recognition. pp.
  6450--6459 (2018)

\bibitem{vondrick2016anticipating}
Vondrick, C., Pirsiavash, H., Torralba, A.: Anticipating visual representations
  from unlabeled video. In: Proceedings of the IEEE conference on computer
  vision and pattern recognition. pp. 98--106 (2016)

\bibitem{wang2021unsupervised}
Wang, G., Zhou, Y., Luo, C., Xie, W., Zeng, W., Xiong, Z.: Unsupervised visual
  representation learning by tracking patches in video. In: Proceedings of the
  IEEE/CVF Conference on Computer Vision and Pattern Recognition. pp.
  2563--2572 (2021)

\bibitem{wang2020self}
Wang, J., Jiao, J., Liu, Y.H.: Self-supervised video representation learning by
  pace prediction. In: European conference on computer vision. pp. 504--521.
  Springer (2020)

\bibitem{wang2015unsupervised}
Wang, X., Gupta, A.: Unsupervised learning of visual representations using
  videos. In: Proceedings of the IEEE international conference on computer
  vision. pp. 2794--2802 (2015)

\bibitem{xu2019self}
Xu, D., Xiao, J., Zhao, Z., Shao, J., Xie, D., Zhuang, Y.: Self-supervised
  spatiotemporal learning via video clip order prediction. In: Proceedings of
  the IEEE/CVF Conference on Computer Vision and Pattern Recognition. pp.
  10334--10343 (2019)

\bibitem{yao2020video}
Yao, Y., Liu, C., Luo, D., Zhou, Y., Ye, Q.: Video playback rate perception for
  self-supervised spatio-temporal representation learning. In: Proceedings of
  the IEEE/CVF Conference on Computer Vision and Pattern Recognition. pp.
  6548--6557 (2020)

\bibitem{zhang2020federated}
Zhang, F., Kuang, K., You, Z., Shen, T., Xiao, J., Zhang, Y., Wu, C., Zhuang,
  Y., Li, X.: Federated unsupervised representation learning. arXiv preprint
  arXiv:2010.08982  (2020)

\bibitem{zhao2017temporal}
Zhao, Y., Xiong, Y., Wang, L., Wu, Z., Tang, X., Lin, D.: Temporal action
  detection with structured segment networks. In: Proceedings of the IEEE
  International Conference on Computer Vision. pp. 2914--2923 (2017)

\bibitem{zhuang2021collaborative}
Zhuang, W., Gan, X., Wen, Y., Zhang, S., Yi, S.: Collaborative unsupervised
  visual representation learning from decentralized data. In: Proceedings of
  the IEEE/CVF International Conference on Computer Vision. pp. 4912--4921
  (2021)

\end{thebibliography}
